\documentclass[letterpaper, 10 pt, conference]{ieeeconf}

\IEEEoverridecommandlockouts
\usepackage{graphicx}
\usepackage{amsmath,amssymb}
\usepackage{booktabs}
\usepackage{multirow}
\usepackage{array}
\usepackage{siunitx}
\usepackage{url}
\usepackage{balance}
\usepackage[utf8]{inputenc}

\title{\LARGE \bf
CriticalScenes: Scalable Photorealistic Critical-Scenario Generation
for Closed-Loop Driving Adaptation
}

\author{
Alexandros Theocharous \qquad Holger Caesar \qquad Ziqi Zhang\\[6pt]
{\small Delft University of Technology}\\
{\small Delft, The Netherlands}\\[2pt]
{\small\ttfamily
a.theocharous@student.tudelft.nl,
H.Caesar@tudelft.nl,
Z.Zhang-16@tudelft.nl}
}

\IEEEaftertitletext{%
\centering
\includegraphics[width=0.76\textwidth]{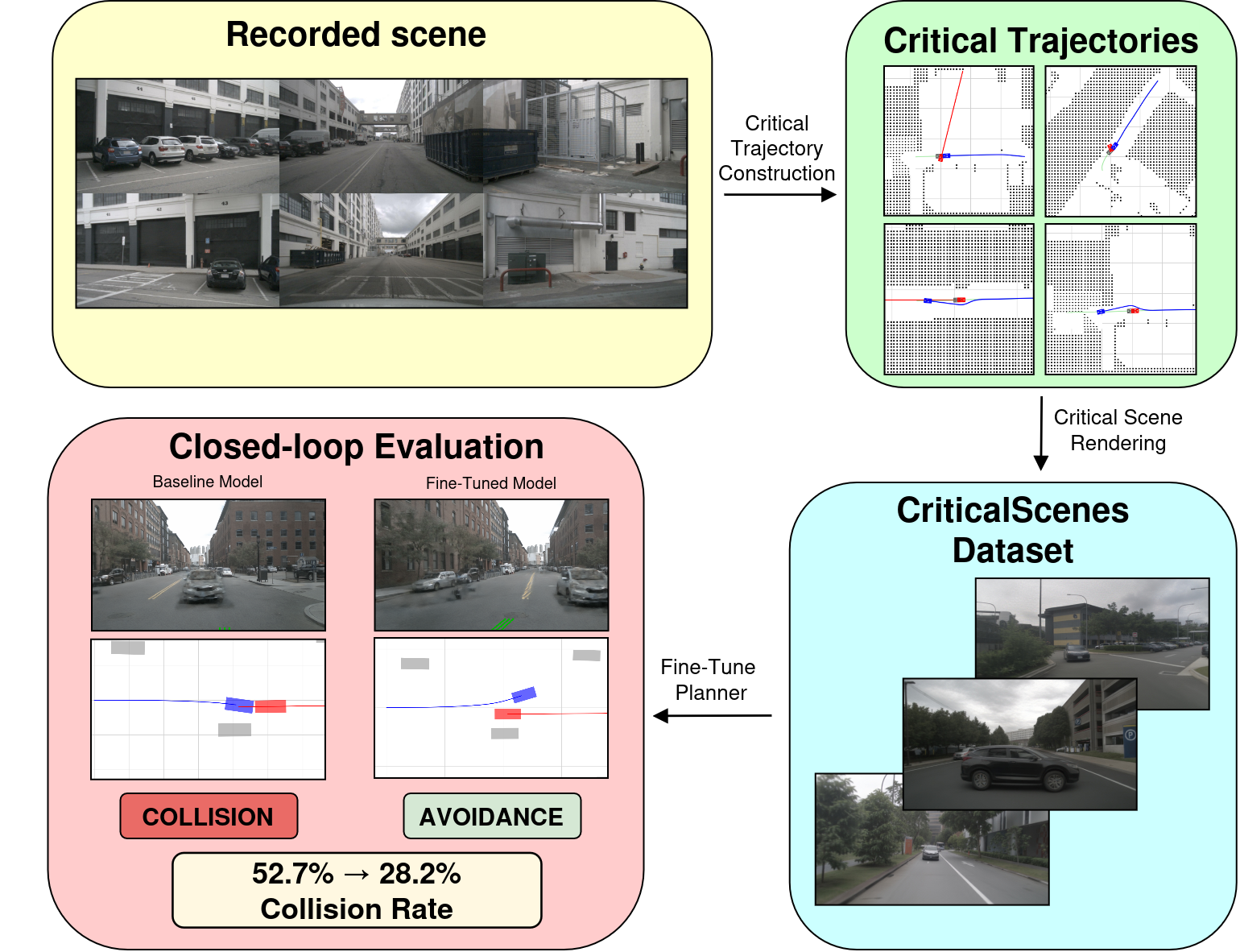}\\[0.8mm]
\parbox{0.96\textwidth}{%
\footnotesize
\textbf{Fig. 1. CriticalScenes concept and evaluation.}
CriticalScenes transforms ordinary recorded driving data into targeted safety-critical training demonstrations for planner adaptation. The adapted planner is evaluated independently in NeuroNCAP, where SparseDrive reduces its collision rate from \textbf{52.7\% to 28.2\%}.
}\\[1.5mm]
}

\begin{document}
\maketitle
\thispagestyle{empty}
\pagestyle{empty}

\renewcommand{\footnoterule}{%
  \kern -3pt
  \hrule width 0.4\columnwidth height 0.4pt
  \kern 5pt
}

{\renewcommand{\thefootnote}{}%
\footnotetext{%
This work has been submitted to the IEEE for possible
publication. Copyright may be transferred without notice,
after which this version may no longer be accessible.}%
}

\setcounter{figure}{1}

\begin{abstract}
Safety-critical interactions are rare in recorded driving data, leaving end-to-end driving planners with limited exposure to evasive behaviour during training. This lack of critical supervision can lead to unsafe behaviour and, in rare instances, collisions. To address this gap, we introduce CriticalScenes, a reconstruction-based framework that transforms ordinary recorded driving scenes into photorealistic safety-critical interactions with corresponding collision-avoidance supervision. Across 53 reconstructed source scenes, 979 of 1,060 attempted interactions yield feasible threat trajectories (92.4\%), while the inserted threat actors remain detectable in more than 95\% of non-occluded observations within 5--35\,m. Experiments with two structurally different end-to-end planners show that the generated supervision improves critical-scenario planning, while closed-loop transfer varies across planner architectures. SparseDrive achieves a 46.5\% relative reduction in collision rate on an independent photorealistic closed-loop benchmark, whereas UniAD shows only a modest closed-loop improvement despite substantial open-loop gains. The results further highlight the importance of closed-loop evaluation, as improvements in open-loop trajectory prediction do not necessarily translate into equivalent safety gains.
\end{abstract}

\section{Introduction}

End-to-end autonomous-driving planners increasingly integrate perception, prediction, and planning within unified learning frameworks and have achieved strong performance on large-scale real-world benchmarks~\cite{chen2024endtoendautonomousdrivingchallenges,hu2023planningorientedautonomousdriving,sun2024sparsedriveendtoendautonomousdriving}. However, these systems are predominantly trained from recorded human driving, in which imminent collisions and emergency avoidance manoeuvres occur only rarely~\cite{westhofen2023criticality,song2023critical}. Consequently, strong performance on conventional open-loop benchmarks does not necessarily imply that a planner has learned how to react when rapid evasive steering or emergency braking is required. Closed-loop safety benchmarks such as NeuroNCAP expose this limitation by placing learned planners in photorealistic collision-course interactions and directly evaluating whether the resulting collision is avoided~\cite{ljungbergh2024neuroncap}. Such benchmarks reveal safety-critical failure modes, but identifying these failures does not by itself provide the targeted training demonstrations required to correct them.

Synthetic data generation offers a way to provide this missing supervision without deliberately collecting dangerous interactions in the real world. Classical simulators such as CARLA provide precise control over traffic actors and scenario geometry~\cite{dosovitskiy2017carlaopenurbandriving}, while generative video and world-model approaches can produce increasingly realistic and diverse driving observations~\cite{li2023drivingdiffusionlayoutguidedmultiviewdriving,wen2023panaceapanoramiccontrollablevideo,hu2023gaia1generativeworldmodel}. However, maintaining temporal and multi-view consistency while retaining precise control over individual actors remains challenging. Reconstruction-based approaches provide an attractive alternative by modifying recorded real-world environments while preserving much of their original appearance and geometry~\cite{yang2023unisimneuralclosedloopsensor,tonderski2024neuradneuralrenderingautonomous,yan2024streetgaussiansmodelingdynamic}. Nevertheless, automatically converting real-world driving recordings into controlled critical interactions paired with collision-avoidance supervision remains an open challenge.

To address this gap, we introduce CriticalScenes, a reconstruction-based framework for generating photorealistic, planner-ready safety-critical demonstrations from ordinary driving data. An overview of the proposed framework and its use for closed-loop planner adaptation is shown in Fig.~1. Starting from editable 3D Gaussian Splatting reconstructions~\cite{yan2024streetgaussiansmodelingdynamic}, CriticalScenes transforms ordinary recorded driving scenes into photorealistic safety-critical interactions and automatically generates collision-avoidance supervision for planner training. The resulting photorealistic multi-view sequences preserve the surrounding real-world scene context while providing the observations, annotations, and planning supervision required for end-to-end planner training and evaluation. In this way, ordinary driving recordings can be systematically repurposed into targeted collision-avoidance demonstrations that are rarely represented in real-world data.

We evaluate CriticalScenes using two structurally different end-to-end planners, UniAD~\cite{hu2023planningorientedautonomousdriving} and SparseDrive~\cite{sun2024sparsedriveendtoendautonomousdriving}. Both planners are fine-tuned using the generated collision-avoidance supervision and evaluated on held-out critical scenarios, but their closed-loop transfer differs substantially. UniAD achieves only a modest aggregate improvement on NeuroNCAP, whereas SparseDrive achieves a $46.5\%$ relative reduction in collision rate on the independent closed-loop benchmark~\cite{ljungbergh2024neuroncap}. These results show that CriticalScenes can expose safety-critical planner failures and provide targeted supervision for improving them, while also highlighting that the effectiveness of this supervision depends on the planner and its adaptation mechanism. Building on these observations, our work makes the following three contributions:

\begin{itemize}
\item \textbf{Automated critical-interaction generation.} We introduce a method for transforming ordinary recorded driving scenes into controlled collision-course interactions by automatically selecting conflict locations and suitable threat actors, and generating temporally synchronised threat trajectories together with feasible steering- or braking-based ego responses.

\item \textbf{Photorealistic planner-ready data generation.} We realise the generated interactions in editable 3D Gaussian reconstructions and combine novel-view rendering, diffusion-based restoration, quality control, and dataset conversion to produce multi-view camera observations and annotations suitable for end-to-end planner training and evaluation.

\item \textbf{Planner adaptation and closed-loop validation.} We evaluate CriticalScenes with UniAD and SparseDrive, showing substantial closed-loop collision reduction for SparseDrive while revealing markedly different transfer behaviour across planner architectures.
\end{itemize}

\section{Related Work}

\subsection{End-to-End Autonomous Driving}

Structured end-to-end planners such as UniAD~\cite{hu2023planningorientedautonomousdriving}, VAD~\cite{jiang2023vadvectorizedscenerepresentation}, and SparseDrive~\cite{sun2024sparsedriveendtoendautonomousdriving} jointly optimise perception, prediction, and planning while retaining intermediate scene representations. Although these methods perform strongly on datasets such as nuScenes, their training data predominantly consists of naturalistic driving and therefore contains few interactions requiring emergency avoidance. This motivates the generation of realistic critical scenarios together with corrective driving supervision.

\subsection{Critical-Scenario Data Generation}

Simulation provides a direct mechanism for generating safety-critical interactions. Platforms such as CARLA~\cite{dosovitskiy2017carlaopenurbandriving} and MetaDrive~\cite{li2022metadrivecomposingdiversedriving} offer explicit control over traffic actors, road geometry, and environmental conditions, enabling the automatic generation of large numbers of scenarios. However, the visual and behavioural domain gap between synthetic environments and real-world driving data can limit the direct use of these environments for training perception-dependent end-to-end planners.

Prior work has also explicitly targeted the generation of safety-critical scenarios. Methods such as AdvSim~\cite{wang2023advsimgeneratingsafetycriticalscenarios}, KING~\cite{hanselmann2022kinggeneratingsafetycriticaldriving}, and Learning to Collide~\cite{ding2020learningcollideadaptivesafetycritical} perturb traffic participants or optimise scenario parameters to expose failures in autonomous-driving policies. These approaches provide systematic mechanisms for generating adversarial interactions, but generally operate through simulation, trajectory perturbation, or abstract scene representations rather than producing photorealistic planner-training data grounded in reconstructed real-world environments.

Recent generative approaches attempt to reduce this gap by synthesising realistic driving observations using diffusion models or learned world models. Methods such as MagicDrive~\cite{gao2024magicdrivestreetviewgeneration}, Panacea~\cite{wen2023panaceapanoramiccontrollablevideo}, ScenarioControl~\cite{gao2026scenariocontrolvisionlanguagecontrollablevectorized}, and GAIA-1~\cite{hu2023gaia1generativeworldmodel} can generate diverse driving scenes conditioned on semantic or geometric information. While these methods substantially improve visual realism and diversity, precise control over the geometry and timing of individual traffic interactions remains challenging. Generative models can additionally introduce inconsistencies or hallucinations that are undesirable when the generated data must provide accurate geometric supervision for trajectory planning.

An alternative direction reconstructs real driving scenes and subsequently modifies or re-renders them. Neural rendering and 3D Gaussian Splatting methods such as NeuRAD~\cite{tonderski2024neuradneuralrenderingautonomous}, Street Gaussians~\cite{yan2024streetgaussiansmodelingdynamic}, and HUGSIM~\cite{zhou2024hugsimrealtimephotorealisticclosedloop} preserve the appearance of recorded environments while enabling novel viewpoints and scene manipulation. This makes reconstruction-based simulation particularly attractive for planner training because generated observations remain close to the real sensor domain.

Recent work has begun to use reconstructed scenes directly for improving driving policies. Ge et al.~\cite{ge2025unravelingeffectssyntheticdata} introduce SceneCrafter, which generates diverse interactive driving data for improving end-to-end planner generalisation, but does not explicitly target collision-course interactions or pair them with corresponding corrective ego trajectories. Concurrent work, World Engine~\cite{li2026worldengine}, identifies failure-prone events in real driving logs, reconstructs them as photorealistic interactive environments, and generates behavioural variations for reinforcement-based post-training. However, its scenario generation remains grounded in long-tail events already present in the recorded logs. In contrast, CriticalScenes constructs collision-course interactions from ordinary recorded scenes and directly pairs each generated threat with a corresponding collision-avoidance trajectory, producing explicit planner-ready supervision without requiring that the critical event be observed in the original data.

\subsection{Closed-Loop Safety Evaluation and Adaptation}

Autonomous driving planners are commonly evaluated using open-loop metrics that compare predicted trajectories with recorded expert trajectories. Such metrics are convenient but cannot capture the feedback between the planner's actions and the evolving scene. A trajectory that appears reasonable against logged ground truth may therefore behave differently when repeatedly replanned in an interactive environment.

Closed-loop benchmarks address this limitation by allowing the planner's predictions to influence subsequent observations and vehicle motion. NeuroNCAP~\cite{ljungbergh2024neuroncap} evaluates camera-based planners in reconstructed nuScenes environments and introduces controlled perturbations designed to expose unsafe behaviour. Such evaluations demonstrate that strong open-loop performance does not necessarily translate into robust behaviour under safety-critical closed-loop interactions.

These findings motivate a complementary use of reconstructed environments: not only to expose planner failures, but also to generate supervision that can correct them. Taken together, these studies demonstrate that reconstructed environments can support both safety-critical scenario generation and policy improvement. Our work addresses the complementary problem of generating such supervision from ordinary driving recordings: CriticalScenes explicitly constructs collision-course interactions, along with corresponding collision-avoidance supervision, and uses the resulting photorealistic demonstrations for planner adaptation and independent closed-loop evaluation.

\section{Methodology}

\subsection{Framework Overview}

CriticalScenes formulates critical-scenario generation as the construction of an interaction satisfying three requirements: the threat must be kinematically feasible and conflict with the nominal ego motion, the manipulated actor must remain reliably renderable from the required viewpoints, and the interaction must admit a feasible collision-avoidance response. Given a reconstructed driving scene, the framework searches for configurations that satisfy these constraints and converts the resulting interaction and collision-avoidance trajectory into photorealistic, planner-ready supervision. The same reconstructed scene can therefore support multiple independently generated critical interactions without requiring a new environment to be captured or reconstructed for each scenario.

\begin{figure*}[t]
    \centering
    \includegraphics[width=\textwidth]{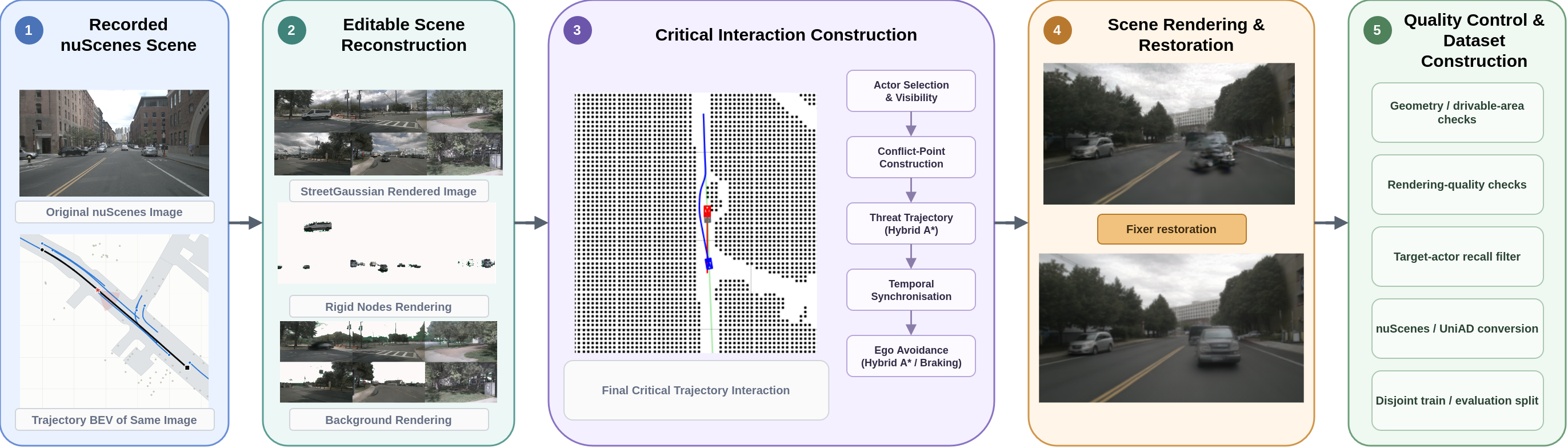}
    \caption{\textbf{Detailed CriticalScenes generation pipeline.} An ordinary recorded scene is reconstructed into an editable representation, after which the framework selects a suitable threat actor and conflict location, constructs the collision-course interaction and collision-avoidance trajectory, renders and restores the modified observations, and applies quality control before conversion to planner-ready training data.}
    \label{fig:pipeline}
\end{figure*}

\subsection{Editable Scene Reconstruction and Threat Selection}

The framework assumes an object-centric reconstruction in which the static environment and foreground traffic participants can be manipulated independently. In our implementation, nuScenes scenes are reconstructed using Street Gaussians~\cite{yan2024streetgaussiansmodelingdynamic}, which represent reconstructed vehicles as individual rigid objects separated from the static Gaussian background. This allows a selected vehicle to be repositioned along a generated trajectory while preserving the remainder of the recorded environment and compatible surrounding traffic.

A critical interaction is defined around a conflict pose $q_c=(x_c,y_c,\psi_c)$ sampled along the recorded future ego trajectory. Candidate threat poses $q_s=(x_s,y_s,\psi_s)$ are sampled from the connected drivable region and retained only when a kinematically feasible trajectory can reach the conflict region within the available interaction time. Candidates are further constrained to approach the ego trajectory from a sufficiently distinct direction, excluding predominantly rear-end configurations that do not require an active evasive response.

Geometric feasibility alone is insufficient because repositioning an actor may expose viewpoints that were poorly observed during reconstruction. We therefore associate each reconstructed actor with a viewpoint-coverage representation derived from its original observations and evaluate its compatibility with the viewpoints required by the candidate interaction. Actors with insufficient support are discarded, while feasible candidates are ranked according to viewpoint compatibility and observation-distance similarity. This biases scenario generation toward configurations that remain reliably renderable after actor repositioning.

\subsection{Critical-Interaction and Safe-Response Generation}

Given the conflict and starting poses, a collision-course threat trajectory is generated using Hybrid A*~\cite{dolgov2008practical}. Hybrid A* provides a kinematically feasible path over the local drivable region while accounting for the non-holonomic motion of a car-like vehicle. Alternative terminal orientations are considered to produce crossing and head-on interactions, and the lowest-cost valid forward trajectory is retained. If no feasible moving trajectory can be found, the reconstructed actor can instead be retained as a stationary obstacle. Rear-end configurations are excluded because the objective is to create interactions requiring an active ego response.

For moving threats, the generated spatial trajectory is temporally synchronised with the recorded ego motion such that the threat and nominal ego reach the conflict region at approximately the same time. Consequently, the original ego trajectory represents the unsafe nominal behaviour; executing it without intervention results in a collision with the generated threat.

A corresponding collision-avoidance trajectory is generated as planning supervision using either lateral avoidance or longitudinal braking. For lateral avoidance, Hybrid A* constructs a trajectory that departs from the nominal route before the conflict and rejoins it after the interaction, thereby completing an evasive manoeuvre. When the interaction can instead be resolved by stopping, a braking response modifies the longitudinal progression along the recorded route such that the ego remains separated from the threat by a spatial safety margin. Moving head-on threats require lateral avoidance and are rejected when no feasible steering response exists.

This process produces three scenario types used in our experiments: crossing, head-on swerve, and head-on brake. Each rendered scenario is directly paired with a corresponding collision-avoidance trajectory for planner training.

\begin{figure*}[t]
    \centering
    \includegraphics[width=\textwidth]{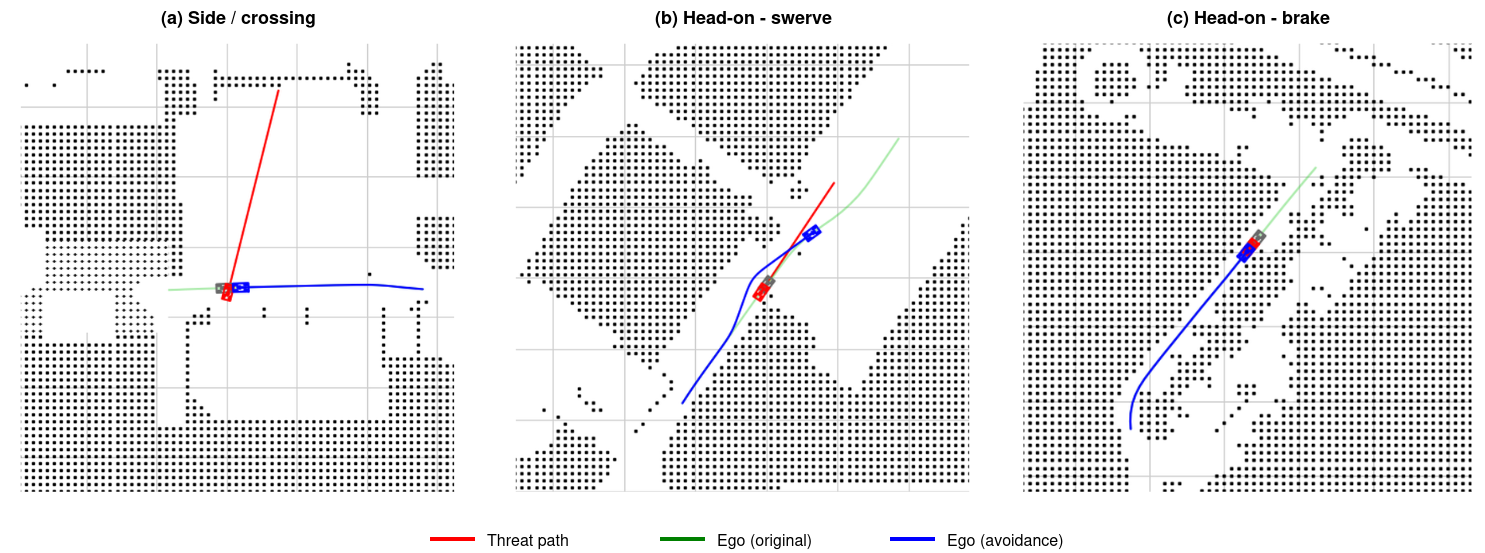}
    \caption{Representative crossing, head-on swerve, and head-on brake scenarios generated by CriticalScenes. Red denotes the generated collision-course threat trajectory or stationary threat location, green the nominal recorded ego trajectory, and blue the generated collision-avoidance trajectory. Swerve responses rejoin the nominal route after passing the conflict region.}
    \label{fig:critical_interactions}
\end{figure*}

\subsection{Rendering, Restoration, and Dataset Construction}

The generated trajectories are applied to the editable reconstruction by moving the selected threat actor along its collision-course trajectory and rendering the cameras from the generated ego poses. Other reconstructed traffic participants retain their recorded motion as long as it remains compatible with the modified interaction, preserving the surrounding traffic context rather than constructing the scenario from an otherwise empty environment.

\begin{figure}[t]
    \centering
    \includegraphics[width=\columnwidth]{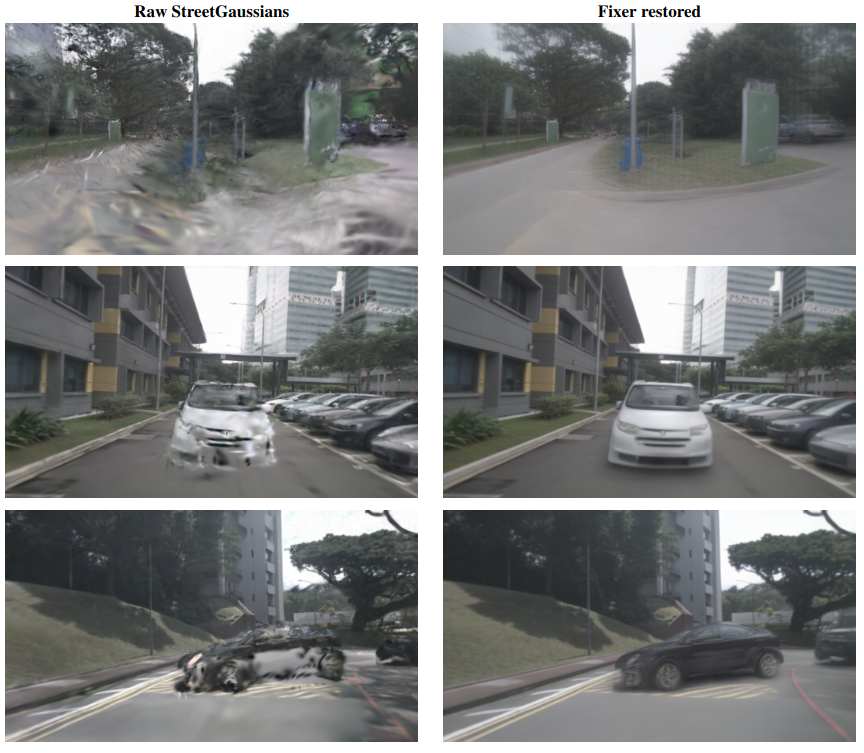}
    \caption{Representative examples of the rendering and restoration stage. Raw Street Gaussians novel-view renderings can contain artefacts when the generated ego or actor poses expose poorly observed regions; Fixer reduces these artefacts before the observations are used for planner training.}
    \label{fig:fixer}
\end{figure}

Modified ego and actor poses can expose viewpoints that were not sufficiently observed during the original reconstruction, leading to incomplete geometry, missing image regions, or malformed actor appearance. The raw novel-view renderings are therefore processed using Fixer~\cite{wu2025difix3dimproving3dreconstructions}, which operates on the rendered observations without altering the generated trajectories or reconstructed scene geometry. Representative examples of the visual improvement are shown in Fig.~\ref{fig:fixer}.

Generated scenarios subsequently undergo geometric and perceptual quality control. Invalid trajectories and interactions with incompatible scene geometry are rejected, while post-render target-actor detection is used to verify that the manipulated vehicle remains perceptually usable by the downstream planner during the critical interaction. Severe reconstruction or restoration failures that cannot yet be reliably captured by automatic criteria are removed through visual inspection.

Accepted scenarios are converted into a planner-compatible nuScenes representation containing multi-view camera observations, ego poses and motion, 3D actor annotations, future trajectories, map information, and planning supervision. All variants generated from the same reconstructed source scene are assigned to the same data split so that closely related versions of an environment do not appear in both training and evaluation.

\subsection{Planner Adaptation}

CriticalScenes is independent of a particular planner architecture: the generated collision-avoidance trajectories use the same coordinate and temporal representation as ordinary planning supervision and can therefore be used to adapt existing end-to-end planners. We evaluate this using two structurally different planners, UniAD~\cite{hu2023planningorientedautonomousdriving} and SparseDrive~\cite{sun2024sparsedriveendtoendautonomousdriving}.

For UniAD, the pretrained perception stack is kept fixed and adaptation is restricted to the downstream motion and planning components. CriticalScenes scenarios are then used as additional supervision for learning collision-avoidance behaviour.

SparseDrive represents planning using a trajectory-anchor vocabulary conditioned on three navigation instructions---left, straight, and right---with six anchors per instruction, for a total of 18 anchors. To introduce evasive trajectory modes, additional anchors are derived directly from the generated CriticalScenes trajectories. For each verified avoidance scenario, the six-step future-planning window at maximum lateral deviation is extracted using the same coordinate transformation used in SparseDrive’s real-data conversion pipeline.
The extracted windows are grouped by navigation instruction and clustered independently using K-means with $K=3$. The resulting three critical centroids are appended to the original six anchors for each instruction, expanding the vocabulary from 18 to 27 anchors overall. The expanded planner is then adapted using CriticalScenes supervision; Section~IV separately evaluates the effects of vocabulary expansion and fine-tuning.

\section{Experiments}

\subsection{Experimental Setup}

We evaluate CriticalScenes from three complementary perspectives: the quality and scalability of the generated scenarios, the effect of the generated supervision on planner behaviour, and the transfer of the learned behaviour to independent closed-loop safety evaluation. The generation pipeline is applied to 53 reconstructed nuScenes source scenes, producing 188 scenarios after trajectory, rendering, and quality-control filtering. From these, 100 scenarios are used for planner adaptation and 20 critical scenarios from source scenes not used for fine-tuning are reserved for evaluation. The training set contains 25 crossing, 39 head-on swerve, and 36 head-on brake scenarios, while the evaluation set contains 5, 10, and 5 scenarios of each type, respectively.

For scenario generation, conflict poses are sampled 5--15\,s ahead along the recorded ego trajectory, and moving threats are generated within a 15--30\, km/h urban-speed range. Actor viewpoint coverage is discretised into eight relative viewing directions. Steering responses are initiated approximately 25\,m before the conflict, while braking responses use a 15\,m reaction threshold.

For UniAD~\cite{hu2023planningorientedautonomousdriving}, fine-tuning is initialised from the public Stage-2 checkpoint and performed for five epochs using AdamW with a learning rate of $2\times10^{-5}$. Training combines 100 randomly selected real nuScenes scenes (4,031 samples) with 100 quality-filtered CriticalScenes entries (3,789 frames spanning 39 reconstructed source scenes). The perception stack remains frozen throughout fine-tuning: the image backbone, image neck, and BEV encoder are hard-frozen, while the detection, segmentation, and temporal query/memory components use zero learning-rate multipliers. Only the motion and planning components are updated.

SparseDrive~\cite{sun2024sparsedriveendtoendautonomousdriving} is fine-tuned for five epochs using AdamW with the same learning rate of $2\times10^{-5}$. Its training mixture contains the same 100 real nuScenes scenes together with 100 generated critical scenarios, corresponding to 3,789 synthetic frames. The image backbone, image neck, detection head, and map head are frozen, while the motion/planning head is adapted. For SparseDrive, closed-loop performance peaked at epoch 2 and degraded with further fine-tuning; we therefore use the epoch-2 checkpoint for all reported adapted results.

Open-loop evaluation is performed on the 20 critical evaluation scenarios and the standard nuScenes validation set, allowing adaptation for critical scenarios and retention of nominal driving to be assessed separately. Closed-loop generalisation is evaluated using NeuroNCAP~\cite{ljungbergh2024neuroncap}, whose NeuRAD-based reconstruction pipeline is independent of the Street Gaussians environment used to generate CriticalScenes. The NeuroNCAP evaluations use the default 100 seeds per scenario combination, corresponding to 2,000 closed-loop runs for each evaluated model, providing a substantially larger evaluation of closed-loop behaviour than the targeted 20-scenario open-loop critical set.

\subsection{Scenario Generation Quality and Perceptual Usability}

We first examine whether CriticalScenes can generate critical interactions reliably and whether the resulting observations remain suitable for downstream planner training. In particular, we evaluate both the efficiency of the reconstruction-based generation pipeline and the perceptual quality of the manipulated target actor. Table~\ref{tab:generation_quality} summarises generation yield, computational cost, and target-actor recall on the 20 evaluation scenarios. Target recall is evaluated only for non-occluded observations, and a prediction is considered a match when its object centre lies within $2\,\mathrm{m}$ of the ground-truth target centre.

\begin{table}[t]
\centering
\caption{Generation yield and computational cost over all 53 source scenes, and target-actor recall on the 20 CriticalScenes evaluation scenarios. Occluded observations are excluded from recall evaluation.}
\label{tab:generation_quality}
\scriptsize
\setlength{\tabcolsep}{4pt}
\renewcommand{\arraystretch}{1.04}

\begin{tabular}{@{}lr@{}}
\toprule
\multicolumn{2}{c}{\textbf{Generation and scalability}} \\
\midrule
Source scenes & 53 \\
Attempts / feasible & 1,060 / 979 \\
Planning success & 92.4\% \\
Selected from feasible & 259 \\
Accepted after quality control & 188 \\
Post-selection acceptance & 72.6\% \\
Reconstruction / source & 59.2 min median \\
Hybrid A* planning / attempt & $\sim$15 s \\
Street Gaussians rendering / entry & 35.1 s \\
NeuRAD rendering / entry & $\sim$90 s \\
Restoration / entry & $\sim$120 s \\
Averaged processing / accepted scenario & $\sim$19.5 min \\
\bottomrule
\end{tabular}

\vspace{2mm}

\begin{tabular}{@{}lcc@{}}
\toprule
\multicolumn{3}{c}{\textbf{Target-actor recall}} \\
\midrule
\textbf{Distance} & \textbf{SparseDrive} & \textbf{UniAD} \\
\midrule
5--15 m & \textbf{99.5\%} & \textbf{98.0\%} \\
15--25 m & 90.4\% & 90.7\% \\
25--35 m & 71.8\% & 65.0\% \\
5--35 m pooled & \textbf{96.3\%} & \textbf{95.4\%} \\
\bottomrule
\end{tabular}
\end{table}

Of 1,060 attempted interactions, 979 produce feasible threat trajectories, corresponding to a planning success rate of 92.4\%. From these feasible trajectories, diversity-preserving selection retains 259 candidates for rendering and downstream validation. Geometric and perceptual quality control, together with visual inspection, then reduces this set to 188 accepted scenarios, corresponding to a 72.6\% post-selection acceptance rate. Scene reconstruction is the dominant one-time cost, requiring a median of 59.2 minutes per source scene. Once an editable reconstruction is available, an additional generated entry requires approximately 15 seconds for Hybrid A* planning, 35.1 seconds for Street Gaussians rendering, and 120 seconds for restoration, corresponding to approximately 2.8 minutes of marginal processing time. Across the 53 reconstructed source scenes, the 188 accepted scenarios correspond to an average of 3.5 scenarios per reconstruction; averaging the median reconstruction time across these yields a nominal processing cost of approximately 19.5 minutes per accepted scenario. Under the same scenario-rendering workload, NeuRAD requires approximately 90 seconds per entry, making the Street Gaussians rendering stage approximately $2.6\times$ faster.

The increased rendering efficiency does not come at the cost of poor target visibility. Within 5--15\,m, SparseDrive and UniAD achieve target recalls of 99.5\% and 98.0\%, respectively, while the pooled recall over 5--35\,m remains at 96.3\% and 95.4\% on the 20-scene evaluation dataset. For comparison, the same planners achieve pooled target recalls of 93.5\% and 91.5\%, respectively, in the NeuRAD-based NeuroNCAP environment before adaptation. Although the two evaluations use different scenes and should not be interpreted as a direct renderer benchmark, these results indicate that actors inserted into the Street Gaussians reconstructions remain comparably perceptible to those encountered in the established NeuRAD-based closed-loop benchmark. Recall decreases at larger distances, where the target occupies fewer image pixels, but remains high during the close-range safety-critical portion of the interaction.

\subsection{Planner Adaptation and Nominal-Driving Retention}

We evaluate whether the generated collision-avoidance trajectories improve planning outside the training set. Table~\ref{tab:openloop} reports L2 trajectory error and object-box collision at 1, 2, and 3\,s on the 20 CriticalScenes evaluation scenarios from source scenes not used for fine-tuning and the standard nuScenes validation set. For UniAD, the pretrained and adapted checkpoints are compared directly. For SparseDrive, we compare the original 18-anchor planner with the final 27-anchor model adapted using CriticalScenes; the effect of vocabulary expansion alone is evaluated separately in the closed-loop test in Table~\ref{tab:closedloop}.

\begin{table*}[t]
\centering
\caption{Open-loop planning performance before and after CriticalScenes adaptation. CriticalScenes results are pooled over the 20 held-out evaluation scenarios. Lower is better.}
\label{tab:openloop}
\scriptsize
\setlength{\tabcolsep}{3.0pt}
\renewcommand{\arraystretch}{1.04}

\begin{tabular}{@{}llcccccc@{}}
\toprule
\textbf{Planner} &
\textbf{Model} &
\multicolumn{3}{c}{\textbf{L2 [m] $\downarrow$}} &
\multicolumn{3}{c}{\textbf{Obj.-box collision [\%] $\downarrow$}} \\
\cmidrule(lr){3-5}
\cmidrule(lr){6-8}
& & \textbf{1 s} & \textbf{2 s} & \textbf{3 s} & \textbf{1 s} & \textbf{2 s} & \textbf{3 s} \\
\midrule

\multicolumn{8}{l}{\textit{CriticalScenes-20}} \\
UniAD & Base
& 7.18 & 12.38 & 16.77
& 6.42 & 11.59 & 9.64 \\
UniAD & Adapted
& 0.62 & 1.25 & 1.99
& 0.28 & 2.37 & 3.63 \\
SparseDrive & Original (18 anchors)
& 0.63 & 1.24 & 1.97
& 7.38 & 9.56 & 10.82 \\
SparseDrive & Adapted (27 anchors)
& 0.90 & 1.62 & 2.39
& 4.20 & 6.71 & 7.69 \\
\midrule

\multicolumn{8}{l}{\textit{nuScenes validation}} \\
UniAD & Base
& 0.84 & 1.21 & 1.77
& 0.12 & 0.38 & 1.01 \\
UniAD & Adapted
& 0.84 & 1.17 & 1.73
& 0.08 & 0.45 & 0.96 \\
SparseDrive & Original (18 anchors)
& 0.30 & 0.58 & 0.95
& 0.01 & 0.06 & 0.23 \\
SparseDrive & Adapted (27 anchors)
& 0.59 & 1.07 & 1.62
& 0.00 & 0.07 & 0.34 \\
\bottomrule
\end{tabular}
\end{table*}

UniAD exhibits a severe baseline weakness on the critical set but learns the generated responses strongly: at 3\,s, L2 decreases from 16.77\,m to 1.99\,m and object-box collision from 9.64\% to 3.63\%, while its nuScenes performance remains largely unchanged. SparseDrive starts from a substantially stronger critical-scenario baseline and further reduces 3\,s object-box collision from 10.82\% to 7.69\%, despite increasing its L2 error from 1.97\,m to 2.39\,m. Thus, both planners improve their collision-related behaviour in the generated critical scenarios, but agreement with a single reference trajectory is insufficient to determine whether the learned behaviour will transfer safely to closed-loop driving.

We therefore evaluate both planners under repeated closed-loop replanning in Section~IV-D.

\subsection{Closed-Loop Safety Generalisation}

The central evaluation asks whether behaviour learned from CriticalScenes transfers to environments beyond the one used to generate the training data. Table~\ref{tab:closedloop} reports NeuroNCAP collision rates by interaction type together with the overall collision rate and NCAP score. For SparseDrive, we evaluate the original 18-anchor planner, the expanded 27-anchor planner without fine-tuning, and the final adapted 27-anchor model under the same full 2,000-run protocol.

\begin{table}[t]
\centering
\caption{NeuroNCAP closed-loop evaluation. For SparseDrive, \emph{original} denotes the pretrained 18-anchor model, \emph{expanded} the 27-anchor model without fine-tuning, and \emph{adapted} the same 27-anchor model fine-tuned with CriticalScenes. Collision rates (CR, \%) are reported by interaction type and overall. Higher NCAP and lower collision rates are better.}
\label{tab:closedloop}
\scriptsize
\setlength{\tabcolsep}{1.8pt}
\renewcommand{\arraystretch}{1.04}

\begin{tabular}{@{}lccccc@{}}
\toprule
\textbf{Model} &
\textbf{Front.} &
\textbf{Stat.} &
\textbf{Side} &
\textbf{CR $\downarrow$} &
\textbf{NCAP $\uparrow$} \\
\midrule
UniAD base
& \textbf{90.2} & 33.1 & \textbf{76.8} & 66.7 & 1.82 \\
UniAD adapted
& 94.0 & \textbf{23.1} & 78.4 & \textbf{65.2} & \textbf{1.94} \\
\midrule
SparseDrive original (18 anchors)
& 69.0 & 56.3 & 32.8 & 52.7 & 3.00 \\
SparseDrive expanded (27 anchors)
& 74.0 & 61.0 & 36.0 & 57.0 & 2.85 \\
SparseDrive adapted (27 anchors)
& \textbf{50.6} & \textbf{19.7} & \textbf{14.4} & \textbf{28.2} & \textbf{3.96} \\
\bottomrule
\end{tabular}
\end{table}

Despite its large open-loop improvement, UniAD obtains only a modest aggregate closed-loop gain, reducing its NeuroNCAP collision rate from 66.7\% to 65.2\%. Its category-level behaviour is mixed: stationary collisions improve, whereas frontal and side collisions worsen slightly.

SparseDrive transfers substantially more strongly. Relative to the original 18-anchor planner, the final CriticalScenes-adapted 27-anchor model reduces the overall collision rate from 52.7\% to 28.2\%, corresponding to a 46.5\% relative reduction, while improving all three interaction categories. This occurs despite the adapted model exhibiting higher open-loop L2 error than the original planner, further showing that reference-trajectory accuracy does not reliably predict closed-loop safety.

The SparseDrive ablation isolates the effect of generated supervision. Expanding the planner's trajectory vocabulary from 18 to 27 anchors without fine-tuning increases the collision rate from 52.7\% to 57.0\%, whereas training the same expanded model with CriticalScenes reduces it to 28.2\%. Over the same comparison, pooled 5--35\,m target-actor recall in NeuroNCAP increases only modestly from 93.5\% to 96.7\%, indicating that the safety improvement cannot be explained by improved target perception alone.

\section{Discussion and Limitations}

The experiments reveal markedly different transfer behaviour across planner architectures. Both planners improve their collision-related open-loop behaviour after training with CriticalScenes, but UniAD achieves only a modest, interaction-dependent closed-loop gain, whereas SparseDrive improves consistently across all NeuroNCAP interaction types. One possible explanation is the difference in how the two planners represent candidate ego motion. SparseDrive explicitly augments its trajectory-anchor vocabulary with critical manoeuvres derived from CriticalScenes, providing dedicated modes for evasive behaviour before fine-tuning. UniAD, in contrast, retains its original planning representation and must incorporate the new behaviour solely through weight adaptation. This may make the generated supervision easier for SparseDrive to exploit under repeated closed-loop replanning. We do not experimentally isolate this architectural factor; therefore, we treat it as a hypothesis rather than a demonstrated cause of the performance difference.

The SparseDrive ablation further clarifies the role of the generated supervision. Expanding the trajectory vocabulary without additional training worsens closed-loop performance, whereas training the same expanded planner on CriticalScenes yields a substantial safety improvement. The gain, therefore, cannot be attributed solely to additional trajectory capacity. Together with the UniAD results, this shows that useful critical-scenario generation requires not only realistic failure cases and corrective demonstrations but also supervision that translates into robust behaviour under repeated replanning. The modest change in target-actor recall and evaluation in the independent NeuRAD-based NeuroNCAP environment further indicates that the observed gain is not attributable to improved perception or to overfitting to the Street Gaussians renderer.

Several limitations remain. The framework currently considers vehicle-based threats and three interaction-response categories. Hybrid A* generates kinematically feasible responses but does not explicitly model full vehicle dynamics, passenger comfort, or human driving style. Reconstruction and novel-view rendering can also introduce artefacts that still require visual quality control. Finally, the experiments use 53 reconstructed source scenes and two end-to-end planners, while the held-out CriticalScenes open-loop evaluation contains only 20 scenarios. The corresponding open-loop results should therefore be interpreted as a targeted assessment rather than a comprehensive estimate of performance across critical interactions. Future work should broaden scenario and planner diversity, evaluate larger held-out critical sets with uncertainty estimates, incorporate stronger dynamic constraints, and further automate quality validation.

\section{Conclusion}

We introduced CriticalScenes, a reconstruction-based framework that transforms ordinary recorded scenes into photorealistic safety-critical interactions with corresponding collision-avoidance supervision. The framework combines automated threat and safe-response generation, editable-scene rendering, restoration, quality control, and planner-ready dataset construction. Experiments with UniAD and SparseDrive show that the generated supervision improves critical-scenario behaviour but transfers differently across planner architectures. Most notably, the final CriticalScenes-adapted SparseDrive planner reduces NeuroNCAP collision rate from 52.7\% to 28.2\%, a 46.5\% relative reduction, while vocabulary expansion alone worsens safety. Overall, these results demonstrate that ordinary recorded driving scenes can be systematically transformed into targeted photorealistic critical-scenario supervision that improves closed-loop collision avoidance, without requiring the critical interaction to be present in the original recording.

\section*{Acknowledgment}

ChatGPT (OpenAI) was used for language editing and coding assistance during the preparation of this work. The research conceptualisation, methodology, experimental design, analysis, and conclusions were developed by the authors. All AI-assisted text and code were reviewed and verified by the authors, who take full responsibility for the final work.

\balance
\bibliographystyle{IEEEtran}
\bibliography{references}

\end{document}